\documentclass[letterpaper, 10 pt, conference]{ieeeconf}  

\usepackage[utf8]{inputenc} 
\usepackage[T1]{fontenc}    
\usepackage{hyperref}       
\usepackage{url}            
\usepackage{booktabs}       
\usepackage{amsfonts}       
\usepackage{nicefrac}       
\usepackage{microtype}      
\usepackage{xcolor}         
\usepackage{graphicx}
\usepackage{amsmath}
\usepackage{amssymb}
\usepackage{booktabs}
\usepackage{comment}
\usepackage{amsmath,amssymb} 
\usepackage{color}
\usepackage{tabularx}
\usepackage{booktabs}
\usepackage{multirow}
\usepackage{array}
\newcolumntype{H}{>{\setbox0=\hbox\bgroup}c<{\egroup}@{}}
\usepackage{xcolor,colortbl}
\usepackage[linesnumbered,ruled,vlined]{algorithm2e}

\usepackage{algpseudocode}
\usepackage{float}
\usepackage{subcaption}
\usepackage{url}
\usepackage[font=normal,labelfont=bf]{caption}
\usepackage{titlesec}
\usepackage{booktabs, cellspace, multirow}
\definecolor{bleudefrance}{rgb}{0.19, 0.55, 0.91} 
\definecolor{babyblue}{rgb}{0.54, 0.81, 0.94}
\definecolor{brightturquoise}{rgb}{0.3, 0.51, 0.87}

\newcommand{\ours}{CSC-Tracker }
\newcommand{\ie}{i.e., }
\newcommand\notsotiny{\@setfontsize\notsotiny{6}{7}}
\title{\LARGE \bf Multi-Object Tracking by Hierarchical Visual Representations}

\author{Jinkun Cao$^{1}$, Jiangmiao Pang$^{2}$ and Kris Kitani$^{1}$\\
$^1$\small Carnegie Mellon University \; $^2$Shanghai AI Laboratory
}

\begin{document}

\maketitle
\thispagestyle{empty}
\pagestyle{empty}

\begin{abstract}
We propose a new visual hierarchical representation paradigm for multi-object tracking. 
It is more effective to discriminate between objects by attending to objects' \textit{compositional} visual regions and contrasting with the background \textit{contextual} information instead of sticking to only the \textit{semantic} visual cue such as bounding boxes. 
This \textit{compositional-semantic-contextual} hierarchy is flexible to be integrated in different appearance-based multi-object tracking methods.
We also propose an attention-based visual feature module to fuse the hierarchical visual representations.
The proposed method achieves state-of-the-art accuracy and time efficiency among query-based methods on multiple multi-object tracking benchmarks.
\end{abstract}

\section{Introduction}
Discriminative visual representations can help avoid mismatches between different targets in appearance-based association for multi-object tracking.
We propose a new visual representation paradigm by fusing visual information from different spatial regions in a hierarchy. 
We argue that, compared to the common paradigm of only using features from bounding boxes, the proposed hierarchical visual representation is more discriminative and no extra annotations are required.

\begin{figure}
    \centering
    \includegraphics[width=\linewidth]{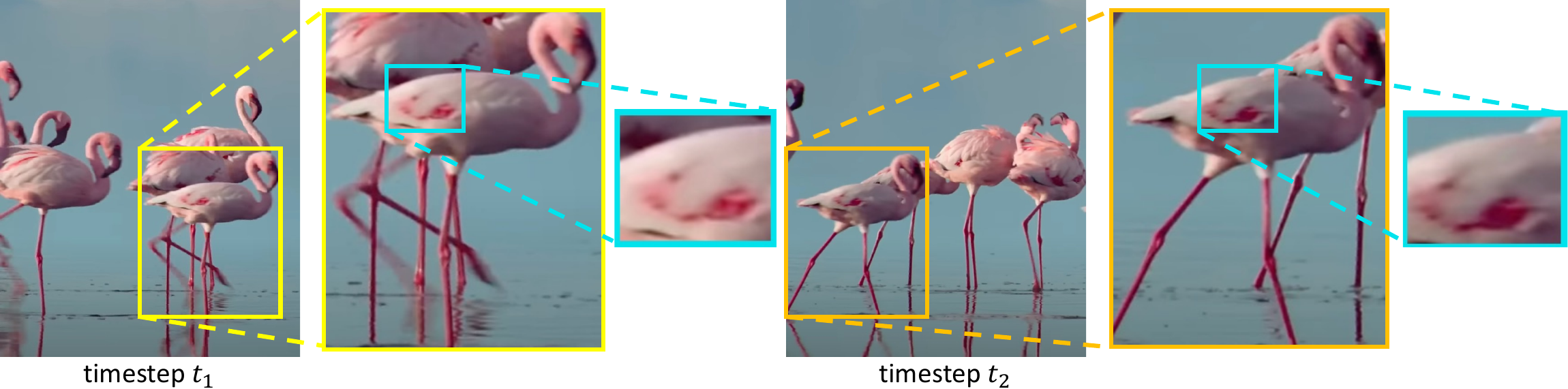}
    \caption{\small With a close look at distinct compositional visual regions, we can recognize certain individuals much more easily.}
    \label{fig:flamingo}
    \vspace{-0.5cm}
\end{figure}

In modern computer vision, we typically use bounding boxes or instance masks to define the area of an object of interest. Because the enclosed pixel area is bonded with a certain object category, such a representation is usually considered as \textit{semantic}.
However, we find that not just the \textit{semantic} cues can make informative representations for visual recognition. We can generate more discriminative visual representations from the other two perspectives to define the existence of an object: \textit{compoistional} and \textit{contextual}.
Compositional cues describe how the parts of a target look like and contrast cues describe how a target looks different from others. 
For example, as shown in Figure~\ref{fig:flamingo}, multiple flamingo individuals are in almost undistinguishable appearance to us. 
But by focusing on the distinguishable parts of certain individuals, such as the shape of the wing red mark, we can easily spot the individual (\textit{compositional}). We can also be more confident in distinguishing instances if we can compare all individuals across timesteps (\textit{contrast}). 
 
We thus build discriminative visual representations from three perspectives: \textit{compositional}, \textit{semantic}, and \textit{contextual}. 
The \textit{semantic} level, such as a tight bounding box or instance segmentation mask, defines the occupancy area of the object with certain visual existence and semantic concept. 
The compositional level suggests the salient visual regions of an object instance, with which, ideally, we can track it even without seeing its full body. 
The \textit{contextual} information helps to highlight a subject via contrast with background pixels and other instances. 
For example, we often have a hard time determining whether two object instances are the same one. However, it is typically easier to determine whether one instance is more likely to be the same one than another. 
Motivated by the insight, we propose to represent an object by a three-level hierarchy, i.e., \textit{Compositional}, \textit{Semantic}, and \textit{Contextual}. 

We adopt the proposed visual hierarchy in video multi-object tracking to avoid the mismatch among different targets.
We find that it is crucial how the representations from levels are leveraged together. The naive way of stacking or concatenating them does not show a significant performance advantage. 
Instead, we propose an attention-based module called CSC-Attention to fuse the features. The core idea of CSC-Attention is to leverage the attention-based mechanism to attend to the salient areas on the target subject body by contrasting to the background pixels close to it. 
Discriminating targets by the fused features, the multi-object tracker we construct is named \ours. 
It leverages global association by a transformer to effectively track objects over time. Through experiments on multiple multi-object tracking datasets, \ours achieves state-of-the-art accuracy among transformer-based methods with better robustness to noise, better time efficiency, and more economic computation requirements.

Our contributions are three-fold. 
First, we propose a visual hierarchy for more discriminative visual representations without additional annotations. 
Second, we propose an attention-based module to leverage the hierarchical features. 
Last, we build a transformer-based tracker with these two innovations and demonstrate its superior accuracy and time efficiency in a pure appearance-based fashion for multi-object tracking.

\section{Related Works}
\noindent\textbf{Deep Visual Representation.}
We typically use a backbone network to extract features from a certain area, such as bounding boxes, as a visual representation for visual perception.
However, the bounding box is noisy as it always contains pixels from the background or other object instances.
For a more fine-grained visual representation, a common way is to use pre-defined regions, such as human head~\cite{headtrack,shao2018crowdhuman} or human joints~\cite{andriluka2018posetrack,poseflow}. 
However, these choices require additional data annotations and specified perception modules. 
Without requiring additional annotations, multi-region CNN~\cite{multiregioncnn} proposes to stack the features from bounding box bins to build a compositional visual representation. 
However, this paradigm can not generate instance-level discriminative representation though it shows effectiveness in semantic-level recognition. 
Moreover, simply stacking features can't emphasize the discriminative visual regions.

\noindent\textbf{Hierarchy Visual Representations.}
The term ``hierarchical visual representations'' has been used indiscriminately for (1) features fused from different resolutions of the same area, such as CNN feature pyramid~\cite{ma2018robust,lin2017feature} and (2) features fused from different pixel areas. Our proposed hierarchical visual representations lie in the second genre. 
Our idea is inspired by David Marr's hierarchical modeling of the human body~\cite{marr2010vision} (\textit{computational}, \textit{algorithmic}, and \textit{implementational})
and the visual cognitive hierarchy~\cite {fodor1988connectionism} (\textit{semantic}, \textit{syntactic}, \textit{physical}). 
Compared to the two visual hierarchies, the three-level hierarchy we propose (\textit{compositional}, \textit{semantic}, \textit{contextual}) is focused on building discriminative visual representations for multi-object tracking. 
Also, in the area of re-identification, some previous works leverage part-based hierarchical features to build visual representation. 
But most of them typically require additional annotations for body parts~\cite{somers2023body}. The way they fuse the features from different regions~\cite{fu2019horizontal} is not effective in multi-object tracking cases where the background noise in the target bounding box area is usually more severe with fast-moving targets and non-static cameras. 

\noindent\textbf{Query-based Multi-Object Tracking.}
Transformer~\cite{vaswani2017attention} is introduced to visual perception~\cite{carion2020end} after its original application in natural language processing. Later, query-based multi-object tracking methods were proposed.
The early methods~\cite{sun2020transtrack,meinhardt2021trackformer} associate objects locally on adjacent time steps. Some recent methods associate targets globally in a video clip~\cite{gtr,zeng2021motr}. GTR~\cite{gtr} removes secondary modules such as positional encoding, making a clean baseline to evaluate feature discriminativeness.
Most recent methods improve performance by gathering information over a long period~\cite{memot,zeng2021motr}.
However, a downside is the high requirement of computation resources, e.g., 8xA100 GPUs~\cite{memot}.  Instead, the improvement of our method comes from the proposed hierarchical representation.  
We demonstrate its state-of-the-art effectiveness and efficiency among query-based methods.

\section{Method}
\label{sec:method}
In this section, we first introduce the overall architecture of \ours. Then we describe the proposed CSC-Attention module to fuse the features from the visual hierarchy. Finally, we elaborate on the training and inference of \ours.

\begin{figure*}
    \centering
    \includegraphics[width=.82\linewidth]{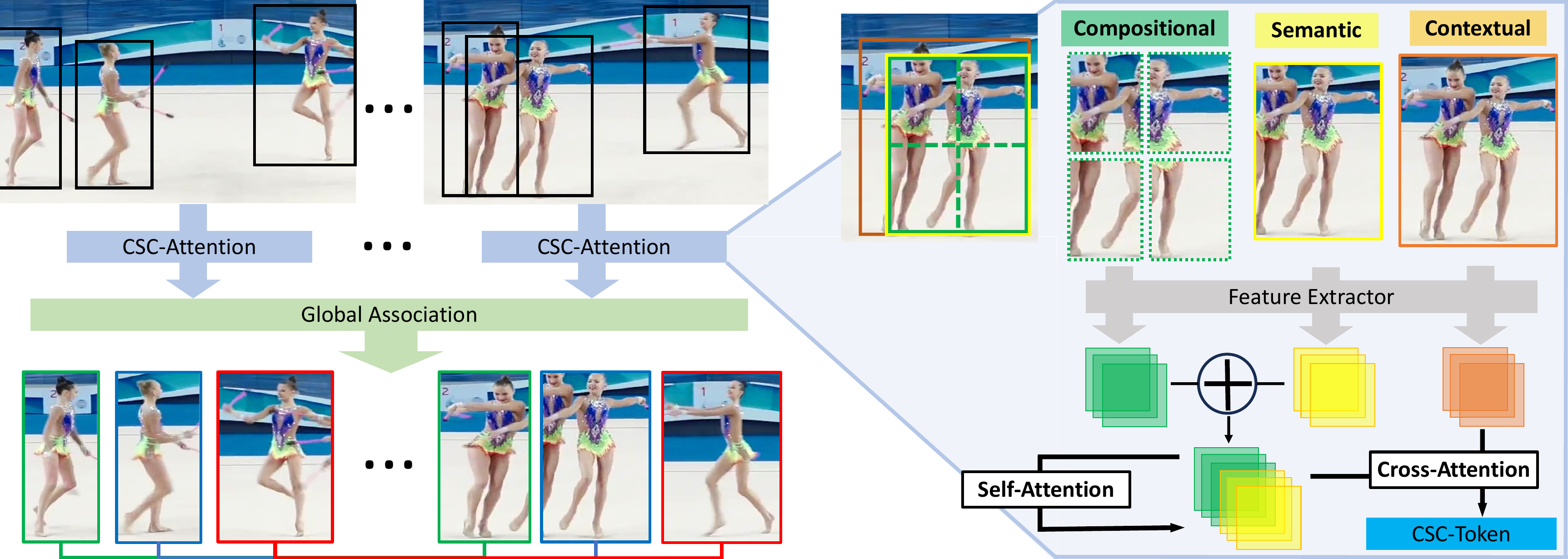}
    \caption{\small The architecture of \ours. The left half illustrates the overall architecture. The right half is the zoomed-in CSC-Attention module. Our contributions are (1) the visual hierarchy for feature extraction and (2) the CSC-Attention module for feature fusion.}
    \label{fig:pipeline}
    \vspace{-0.5cm}
\end{figure*}

\subsection{Overall Architecture}
\label{sec:global_asso}
We follow the spatio-temporal global association paradigm~\cite{vistr,gtr} to build \ours, whose
pipeline is shown in Figure~\ref{fig:pipeline}. Now, we explain the three stages of it. Notations are conditional to a generic time step $t$, which is the last time step where the tracks have been finalized.

\noindent\textbf{Detection and Feature Extraction.} Given a video clip of $T$ frames, \ie $\mathcal{T} =\{t+1,...,t+T\}$, we have the corresponding images $\mathcal{I}=\{I^{t+1}, ..., I^{t+T}\}$. Given a detector, we could derive the detections of the objects of interest on all frames in parallel, noted as $\mathcal{O}=\{O_1,...,O_{N_t}\}$. $N_t$ is the number of detections and $t_i \in \mathcal{T}$ ($1 \leq i \leq N_t$) is the time step where the $i$-th detection, \ie $O_i$, is detected. Then, we extract the features of each detected object by a backbone network.

\noindent\textbf{Token Generation by CSC-Attention.} 
We propose CSC-Attention (to be detailed in the following section) to generate feature tokens. 
By CSC-Attention, we will have the object CSC-tokens $\mathcal{Q}_t^{\text{det}} \in \mathbb{R}^{N_t \times D}$, where $D$ is the feature dimension. If we aim to associate the new-coming detections with existing trajectories, we also need the tokens to represent the existing $M_t$ trajectories, \ie $\mathbf{T}_t^{\text{traj}} = \{Tk_1^{\text{traj}}, Tk_2^{\text{traj}},...,Tk_{M_t}^{\text{traj}}\}$. 
Instead of the resource-intensive iterative query passing~\cite{zeng2021motr} or long-time feature buffering~\cite{memot}, we leverage the CSC-tokens of objects on a trajectory to represent it.
Within a horizon $H$, we represent a trajectory, $Tk_j^{\text{traj}}$, with the token $Q^{\text{traj}}_j \in \mathbb{R}^{H \times D}$ by combining the historical detection CSC-tokens. And all trajectory tokens are $\mathcal{Q}^{\text{traj}}_t = \{Q^{\text{traj}}_1, ..., Q^{\text{traj}}_{M_t}\}$.

\noindent\textbf{Global Association.} By cross-attention, we could get the association score between the set of detections and a trajectory, i.e. $Tk_j^{\text{traj}}$, as $S(Q^{\text{traj}}_j, \mathcal{Q}_t^{\text{det}}) \in \mathbb{R}^{H \times N_t}$. In practice, because we aim to associate between all $M_t$ trajectories and $N_t$ detections, we perform the cross-attention on all object queries and track queries at the same time, namely $S(\mathcal{Q}^{\text{traj}}_t, \mathcal{Q}_t^{\text{det}}) \in \mathbb{R}^{HM_t \times N_t}$. 
By averaging the score on the $H$ steps in the horizon, we get the global association score  $\mathbf{S}^t \in \mathbb{R}^{M_t \times N_t}$. 
Then, we normalize the association scores between a trajectory and objects from the same time step by softmax:
\begin{equation}
    P(\mathbf{M}^t_{j,i} = 1| \mathcal{Q}_t^{\text{det}}, \mathcal{Q}_t^{\text{traj}}) = \frac{\text{exp}(\mathbf{S}^t_{j,i})}{\sum_{k \in \{1,2,...,N_t\}} \mathbf{1}_{[t_k = t_i]} \text{exp}(\mathbf{S}^t_{j,k})},
    \label{eq:asso_score}
\end{equation}
where the binary indicator function $\mathbf{1}_{[t_k = t_i]}$ indicates whether the $i$-th detection and the $k$-th detection are on the same time step. $\mathbf{M}^t \in \mathbb{R}^{(M_t+1) \times N_t}$ is the final global association matrix. 
Its dimension is of ${(M_t+1) \times N_t}$ because each detection can be associated with an ``empty trajectory'' to start a new track. 
The query of the ``empty trajectory'' is represented by a token randomly drawn from a previous unassociated object. 
Also, after the association, unassociated trajectories will be considered absent on the corresponding frames. 
In such a fashion, we can train over a large set of detections and trajectories in parallel and conduct inference online by a sliding window.
We use a uniform form for queries to represent both objects and trajectories. 
Thus, the global association can happen either among detections or between detections and trajectories. 
These two schemes of associations thus are implemented as the same and share all model modules. 
For online inference, we associate detections from the new-coming time step ($T=1$) and existing trajectories. 

\subsection{CSC-Attention}
\label{sec:part_whole_att}
Now, we explain the attention mechanism to fuse the features from the \textit{\textbf{C}ompositional-\textbf{S}emantic-\textbf{C}ontextual} visual hierarchy. We name it CSC-Attention (right-half of Fig.~\ref{fig:pipeline}).

\noindent\textbf{Hierarchy Construction.} 
There are different choices for constructing the hierarchy. 
To have a fair comparison with a close baseline~\cite{multiregioncnn}, we use bounding box bins to represent object parts.
Given a detection $O$, we divide the bounding box into $2 \times 2$ bins (to fit in GPU memory), making a set of body parts as $\mathcal{P} =\{p_1, p_2, p_3, p_4\}$. On the other hand, from a global scope, there are other targets interacting with $O$ which are highly likely to be mismatched in the association stage. We crop the union area enclosing $O$ and all other targets having overlap with it. We note the union area as $U$. Till now, we have derived the triplet $\{\mathcal{P}, O, U\}$ as the raw material for the visual hierarchy.

\noindent\textbf{Feature Fusion.} 
Among the three levels, semantic information is necessary to define a visual boundary. Compositional and contextual cues serve as the enhancement to the final representation's discriminativeness. 
With the extracted regions $\{\mathcal{P}, O, U\}$, we use a shared feature extractor to get their features, i.e. compositional, semantic, and contextual features. 
To fuse the features, we first concatenate the compositional and semantic features. 
Then a self-attention module is applied to help attend to the discriminative regions.
Finally, the contextual features and the self-attention output are processed by a cross-attention module to get the final CSC-tokens.
Before being forwarded to the global association, the tokens would be projected to a uniform dimension of $D$.

\subsection{Training and Inference}
\label{sec:train_inference}
\noindent\textbf{Training.} We train the association module by maximizing the likelihood of associating detections belonging to the same trajectory as in Eq.~\ref{eq:asso_score}. We calculate the association score on all $T$ frames of the sampled video clip simultaneously and globally. 
The objective thus turns to
\begin{equation}
    \max \prod_{q=t+1}^{t+T} P(\mathbf{M}^{t}_{j,\tau_q^j} = 1| \mathcal{Q}_t^{\text{det}}, \mathcal{Q}_t^{\text{traj}}),
\end{equation}
where $\tau_q^j$ is the ground truth index of the detection to be associated with the $j$-th trajectory on the $q$-th time step.
By applying the objective to all trajectories, the training loss is
\begin{equation}
    L_{\text{asso}} = - \sum_{j=1}^{M_t+1} \sum_{q=t+1}^{t+T} \text{log} P(\mathbf{M}^{t}_{j,\tau_q^j} = 1| \mathcal{Q}_t^{\text{det}}, \mathcal{Q}_t^{\text{traj}}).
    \label{eq:asso_loss}
\end{equation}

On the other hand, trajectories can also be absent on some time steps because of occlusion or target disappearance. Therefore, Eq.~\ref{eq:asso_loss} has included the situation of associating a trajectory with no detection, i.e. ``empty''. 
The token for an empty detection is an arbitrary negative sample. We also have a triplet loss to pull away the feature distance between negative pairs compared to that between positive pairs:
\begin{equation}
    \begin{split}
        L_{\text{feat}} = \text{max}(0, \min_{u=1}^{N_P} ||\text{Att}(f(F_{p_u}), f(F_O)) - f(F_O) ||^2 - 
    \\ || \text{Att}(f(F_O), f(F_U^{bg})) -  f(F_O)||^2 + \alpha),
    \end{split}
    \label{eq:feat_loss}
\end{equation}
where $f(\cdot)$ is the shared layers to project CNN features and $N_\mathcal{P}$ is the number of part patches ($N_\mathcal{P}=4$ in our default setting). $\text{Att}(\cdot, \cdot)$ is the operation of cross attention. $\alpha$ is the margin to control the distance between positive and negative pairs. $F_O$ and $F_{p_u}$ ($1 \leq u \leq N_\mathcal{P}$) are the semantic and compositional features. $
F_U^{bg}$ is the features of the background area in the union area $U$. 
We obtain the background features by setting the pixels of $O$ in the area of $U$ to 0 and forward the masked union area into the shared feature encoder $f(\cdot)$. 
We design Eq.~\ref{eq:feat_loss} to encourage (1) the feature encoder to pay more attention to the salient and distinct area on targets while less attention to the background area and (2) the features of the background area in the union box to be discriminative from the foreground object.
Finally, the training objective is 
\begin{equation}
    L = L_{\text{asso}} + L_{\text{feat}} + L_{\text{det}},
\end{equation}
where $L_{\text{det}}$ is an optional detection loss. 

\noindent\textbf{Inference.} We realize online inference by traversing the video with a sliding window of stride 1.
On the first frame, each detection initializes a trajectory. 
By averaging the detection-detection association score alongside a trajectory, we get the detection-trajectory association scores, whose negative value serves as the entries in the cost matrix for the association assignment. 
We adopt Hungarian matching to ensure one-to-one mapping. 
Only when the association score is higher than $\beta=0.3$, the pair can be associated. 
All unassociated detections on the new-coming frames will start new tracks. 

\section{Experiments}
\label{sec:exps}
\begin{table}[!htb]
\centering
\scriptsize
\caption{\footnotesize Results on MOT17 and MOT20 test sets with the private detections (FP and FN reported by $\times 10^4$). }
\setlength{\tabcolsep}{7pt}
\begin{tabular}{ l | p{13px}p{11px}p{11px}p{11px}HHp{6px} p{6px}p{11px}}
\toprule
Tracker & HOTA$\uparrow$ & AssA$\uparrow$ & MOTA$\uparrow$ & IDF1$\uparrow$  & MT & ML &  FP$\downarrow$ & FN$\downarrow$ & IDs$\downarrow$\\
\midrule
\multicolumn{10}{c}{\textbf{MOT-17 Test}} \\
\midrule
FairMOT~\cite{zhang2021fairmot} & 59.3 & 58.0 & 73.7 & 72.3  & & & 2.75 & 11.7 & 3,303 \\
Semi-TCL~\cite{li2021semitcl} & 59.8 & 59.4 & 73.3 & 73.2 & & & 2.29 & 12.5 & 2,790 \\
CSTrack~\cite{cstrack} & 59.3 & 57.9  & 74.9 & 72.6 & &  & 2.38 & 11.4 & 3,567  \\ 
GRTU~\cite{grtu} & 62.0 & 62.1 & 74.9 & 75.0  & & & 3.20 & 10.8 & 1,812 \\ 
QDTrack~\cite{pang2021quasi} & 53.9 & 52.7 & 68.7 & 66.3  & & & 2.66 & 14.7 & 3,378 \\
MAA~\cite{maa} & 62.0 & 60.2  & 79.4 & 75.9 & & & 3.73 & 7.77 & 1,452  \\ 
ReMOT~\cite{yang2021remot} & 59.7 & 57.1 & 77.0 & 72.0 & & & 3.32 & 9.36 & 2,853  \\
PermaTr~\cite{permatrack} & 55.5 & 53.1 & 73.8 & 68.9 & & & 2.90 & 11.5 & 3,699   \\
ByteTrack~\cite{bytetrack} &  63.1  & 62.0 & 80.3 & 77.3 & & & 2.55 &  8.37 & 2,196 \\
DST-Tracker~\cite{dsttrack} & 60.1 & 62.1 & 75.2 & 72.3 & & & 2.42 & 11.0 & 2,729\\ 
UniCorn~\cite{unicorn} & 61.7 & - & 77.2 & 75.5 & & & 5.01 & 7.33 & 5,379\\ 
OC-SORT~\cite{ocsort} & 63.2 & 63.2 & 78.0 & 77.5 & && \underline{1.51} & 10.8 & 1,950\\ 
Deep OC-SORT~\cite{deepocsort} & 64.9 & 65.9 & 79.4 & 80.6 & & & 1.66 & 9.88 & \underline{1,023}\\
MotionTrack~\cite{qin2023motiontrack} & 65.1 & 65.1 & \underline{81.1} & 80.1 & & & 2.38 & 8.17 & 1,140\\ 
SUSHI~\cite{sushi} & \underline{66.5} & \underline{67.8} & \underline{81.1} & \underline{83.1} & & & 3.23 & \underline{7.32} & 1,149 \\ 
\rowcolor{babyblue!20}TransCt~\cite{transcenter} & 54.5 & 49.7  & 73.2 & 62.2  & & & 2.31 & 12.4 & 4,614  \\
\rowcolor{babyblue!20}TransTrk~\cite{sun2020transtrack}  &54.1 & 47.9 & 75.2 & 63.5  && & 5.02 & \textbf{8.64} & 3,603   \\
\rowcolor{babyblue!20}MOTR~\cite{zeng2021motr} &  57.2 & 55.8  &71.9 & 68.4 & & &  \textbf{2.11} & 13.6 & \textbf{2,115}\\
\rowcolor{babyblue!20}TrackFormer~\cite{meinhardt2021trackformer} &  - & -  & 65.0 & 63.9 & & & 7.44 & 12.4 & 3,528  \\
\rowcolor{babyblue!20}GTR~\cite{gtr} &  59.1 & 57.0 & 75.3 & 75.1 & & & 2.68 & 10.9 & 2,859 \\
\rowcolor{babyblue!20} MeMOT~\cite{memot} &  56.9 & 55.2 & 72.5 & 69.0 & & & 3,72 & 11.5 & 2,724  \\
\rowcolor{babyblue!20} \ours  & \textbf{60.8} & \textbf{60.7} & \textbf{75.4} & \textbf{75.7} & & & 2,45 & 10,8 & 2,879 \\
\midrule
\multicolumn{10}{c}{\textbf{MOT-20 Test}}\\
\midrule
FairMOT~\cite{zhang2021fairmot} & 54.6 & 54.7 & 61.8 & 67.3 & 68.8  & \underline{7.6 } & 10.3 & 8.89 & 5,243\\
CSTrack~\cite{cstrack} & 54.0 &  54.0 & 66.6 & 68.6  & 50.4  & 15.5  & 2.54 & 14.4 & 3,196 \\
GSDT~\cite{gsdt} & 53.6  & 52.7 & 67.1 & 67.5  & 53.1  & 13.2  & 3.19 & 13.5 & 3,131 \\
RelationT~\cite{yu2021relationtrack}  & 56.5 & 55.8& 67.2 & 70.5 & 62.2  & 8.9  & 6.11 & 10.5 & 4,243\\
MAA~\cite{maa} & 57.3 & 55.1  & 73.9 & 71.2 & 741 & 153 & 2.49 & 10.9 & 1,331\\
ByteTrack~\cite{bytetrack} & 61.3 & 59.6 & 77.8 & 75.2 & \underline{69.2 } & 9.5  & 2.62 & 8.76 & 1,223 \\
OC-SORT~\cite{ocsort} & 62.1 & 62.0 & 75.5 & 75.9 & & & 1.80 & 10.8 & 913 \\
Deep OC-SORT~\cite{deepocsort} & \underline{63.9} & \underline{65.7} & 75.6 & \underline{79.2} & && \underline{1.69} & 10.8& \underline{779}\\ 
MotionTrack~\cite{qin2023motiontrack} & 62.8 & 61.8 & \underline{78.0} & 76.5 & & & 2.86 &\underline{8.42} & 1,165\\ 
\rowcolor{babyblue!20} TransCt~\cite{transcenter} & 43.5 & 37.0 & 58.5 & 49.6 & 49.4  & 15.5  & 6.42 & 14.6 & 4,695 \\
\rowcolor{babyblue!20} TransTrk~\cite{sun2020transtrack} & 48.5 & 45.2 & 65.0 & 59.4 & 50.1  & 13.4  & \textbf{2.72} & 15.0 & 3,608  \\
\rowcolor{babyblue!20} MeMOT~\cite{memot} & \textbf{54.1} & \textbf{55.0} & 63.7 & \textbf{66.1} & & & 4,79 & 13.8 & \textbf{1,938} \\ 
\rowcolor{babyblue!20} \ours  & 53.0 & 51.1 & \textbf{65.8} & 64.4 & 677 & 157 & 3.64 & \textbf{13.7} & 3,948 \\
\bottomrule
\end{tabular}
\label{table:mot17_20}
\vspace{-0.5cm}
\end{table}

\subsection{Experiment Setups}
\noindent\textbf{Datasets.}
We focus on pedestrian tracking in this paper as it is the most popular scenario and a line of previous works is available for comparison of association accuracy. On some other tracking datasets, such as TAO~\cite{dave2020tao}, tracking faces main difficulties at the detection stage instead of association. This causes uncontrollable noise to evaluate how discriminative the features are.
For valid evaluation of visual representation distinguishness, we select three datasets, i.e., 
MOT17~\cite{milan2016mot16}, MOT20~\cite{dendorfer2020mot20} and  
DanceTrack~\cite{sun2021dancetrack}. DanceTrack has the largest data scale and provides an official validation set.
DanceTrack contains targets mostly in the foreground but with heavy occlusion, complex motion patterns, and similar appearances. On DanceTrack, detection is not considered as the bottleneck and the model ability of appearance discrimination becomes the key for tracking.

\noindent\textbf{Evaluation Metrics.} 
The CLEAR evaluation protocol~\cite{bernardin2008evaluating} is popular for multi-object tracking evaluation but is biased to single-frame association quality~\cite{luiten2021hota}.
MOTA is the main metric of CLEAR~\cite{bernardin2008evaluating} protocol. 
But it is also biased to the detection quality. To provide a more accurate sense of association accuracy, we emphasize the recent HOTA~\cite{luiten2021hota} metric set where the metric is calculated upon the video-level association between ground truth and predictions (by default in the form of bounding boxes). In the set of metrics, AssA emphasizes the association performance, and DetA stresses the detection quality. HOTA is the main metric by considering both detection and association quality. For the result tables, we use \underline{underlined} numbers to indicate the overall best value and \textbf{bold} numbers for the best query-based methods. All query-based methods are listed in \colorbox{babyblue!50}{blue}.

\noindent\textbf{Implementation.} We use ResNet-50~\cite{he2016deep} as the backbone network, which is pretrained on Crowdhuman~\cite{shao2018crowdhuman} dataset first. Though advanced detector~\cite{bytetrack} is demonstrated as a key to boosting tracking performance, we want our contribution to be more from the improvement of the association stage. 
Therefore, on MOT17, we align the implementation with the practice of GTR~\cite{gtr} to use the classic CenterNet~\cite{centernet,centertrack} as the detector to make a fair comparison.
The CenterNet detector is pretrained together with the backbone on Crowdhuman. For the fine-tuning of association modules on MOT17, we use a 1:1 mixture of MOT17-train and Crowdhuman. We fine-tune with only the MOT20-train for evaluation on MOT20. For DanceTrack, we use its official training set as the only training set during finetuning. The image size is set to be 1280 $\times$ 1280 during training. 
The image size is 1560 for the longer edge during the test. During finetuning, the detector head is also finetuned. 
The training iterations are set to be 20k on MOT17/MOT20 and 80k on DanceTrack. We use BiFPN~\cite{tan2020efficientdet} for the feature upsampling. For the implementation of the transformer, we use a stack of two layers of ``Linear + ReLU'' as the projection layers and one-layer encoders and decoders. We use AdamW~\cite{loshchilov2017decoupled16} optimizer for training whose base learning rate is set to be 5e-5. 
The length of the video clip is $T=8$ for training and $T=24$ for inference in a sliding window for a fair comparison with GTR~\cite{gtr}.  We use 4  $\times$ V100 GPUs as the default training device but we will see that even using only one RTX 3090 GPU for training, our method still achieves comparable performance. The training takes 4 hours on MOT17 or MOT20 and 11 hours on DanceTrack. 

\begin{table}[!htb]
\centering
\scriptsize
\caption{\small Benchmarking results on DanceTrack test set. }
\setlength{\tabcolsep}{7pt}
\begin{tabular}{ l |ccccc}
\toprule
Tracker &  HOTA$\uparrow$ & DetA$\uparrow$ & AssA$\uparrow$ & MOTA$\uparrow$ & IDF1$\uparrow$\\
\midrule
CenterTrack~\cite{centertrack} &  41.8 & 78.1 & 22.6 & 86.8 & 35.7 \\
FairMOT~\cite{zhang2021fairmot} & 39.7 & 66.7 & 23.8 & 82.2 & 40.8\\
QDTrack~\cite{pang2021quasi} & 45.7 & 72.1 & 29.2 & 83.0 & 44.8\\
TraDes~\cite{trades}  & 43.3 & 74.5 & 25.4 & 86.2 & 41.2 \\ 
ByteTrack~\cite{bytetrack} & 47.3 & 71.6 & 31.4 & 89.5 & 52.5\\
OC-SORT~\cite{ocsort} & 55.7 & 81.7 & 38.3 & 92.0 & 54.6\\
Deep OC-SORT~\cite{deepocsort} & 61.3 &\underline{82.2} & 45.8 & \underline{92.3} & 61.5\\
DST-Tracker~\cite{dsttrack} & 51.9 & 72.3 & 34.6 &  84.9 & 51.0\\ 
SUSHI~\cite{sushi} & \underline{63.3} & 80.1 & \underline{50.1} & 88.7 & \underline{63.4}\\ 
\rowcolor{babyblue!20} TransTrk\cite{sun2020transtrack} & 45.5 & 75.9 & 27.5 & 88.4 & 45.2\\
\rowcolor{babyblue!20} MOTR~\cite{zeng2021motr} & 54.2 & 73.5 & 40.2 & 79.7 & 51.5 \\
\rowcolor{babyblue!20} GTR~\cite{gtr} & 48.0 & 72.5 & 31.9 & 84.7 & 50.3\\
\rowcolor{babyblue!20} \ours (Ours)  &  \textbf{55.5} & \textbf{77.3} & \textbf{43.1} & \textbf{89.5} & \textbf{54.0}\\
\bottomrule
\end{tabular}
\label{table:dancetrack}
\vspace{-0.5cm}
\end{table}

\subsection{Benchmark Results}
For benchmarking, we only report the performance of online tracking algorithms as offline post-processing~\cite{du2023strongsort,zhang2023motrv2} gives unfair advantages and blurs the discussion about visual representation discriminativeness.
We first benchmark on MOT17 and MOT20 in Table~\ref{table:mot17_20}. On MOT17, \ours achieves the highest HOTA and AssA score among transformer-based methods.
MOT20 is a more challenging dataset with crowded pedestrian flows. Though \ours shows better performance than MeMOT~\cite{memot} on MOT17, its performance is inferior on MOT20. This is probably related to the long-time heavy and frequent occlusion on MOT20. 
To solve this problem, the long temporal buffer of historical object appearance in MeMOT shows effectiveness. 
However, MeMOT requires 8$\times$A100 GPUs for training to support such a long buffering (22 frames v.s. 8 frames by \ours) and uses COCO~\cite{lin2014microsoft} dataset as the additional pretraining data, which makes it not an apple-to-apple comparison.

We also benchmark on DanceTrack-test in Table~\ref{table:dancetrack}. \ours achieves state-of-the-art performance among transformer-based methods. 
Also, \ours shows advanced time efficiency.
For example, training on MOT17 takes MOTR~\cite{zeng2021motr} 2.5 days on 8$\times$V100 GPUs while only 4 hours on 4$\times$V100 GPUs for our proposed method. The inference speed is 6.3FPS for MOTR while 21.3FPS for our method on the same machine (V100 GPU). Compared to GTR~\cite{gtr}, \ours achieves a more significant outperforming on DanceTrack than on MOT17. As other variables and design choices are strictly controlled, it suggests our proposed visual hierarchy representation is more powerful than the naive bounding box features when the occlusion is heavier. 

Given the aforementioned results, we have demonstrated \ours to be the state-of-the-art among transformer-based methods with a lightweight design. 
More importantly, we show that the proposed hierarchical representation is more effective and efficient in discriminatively distinguishing objects.
\ours builds a new baseline for future research in this line of methods. The commonly adopted techniques of query propagation and iteration~\cite{meinhardt2021trackformer,sun2020transtrack,zeng2021motr}, deformable attention~\cite{sun2020transtrack,memot} and long-time feature buffering~\cite{memot} are all compatible to be integrated with \ours. Compared to the overall state-of-the-art methods, such as OC-SORT~\cite{ocsort} and SUSHI~\cite{sushi}, \ours still shows inferior performance. But their performance is reported with a more advanced detector, i.e. YOLOX~\cite{ge2021yolox}. 
This makes a fair comparison hard to present. But still, there is a performance gap between the SOTAs and the transformer-based methods. For inference speed, given detections on MOT17, OC-SORT runs at 300FPS and SUSHI runs at 21FPS while \ours runs at 93FPS. 

\begin{figure}
\centering
\includegraphics[width=.31\linewidth]{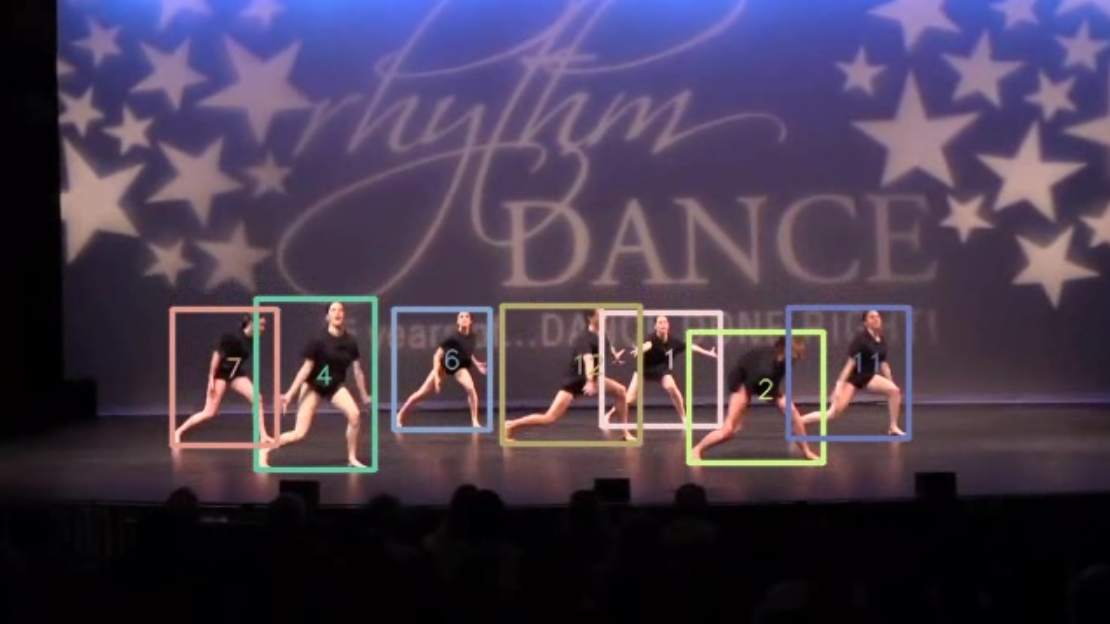}
\includegraphics[width=.31\linewidth]{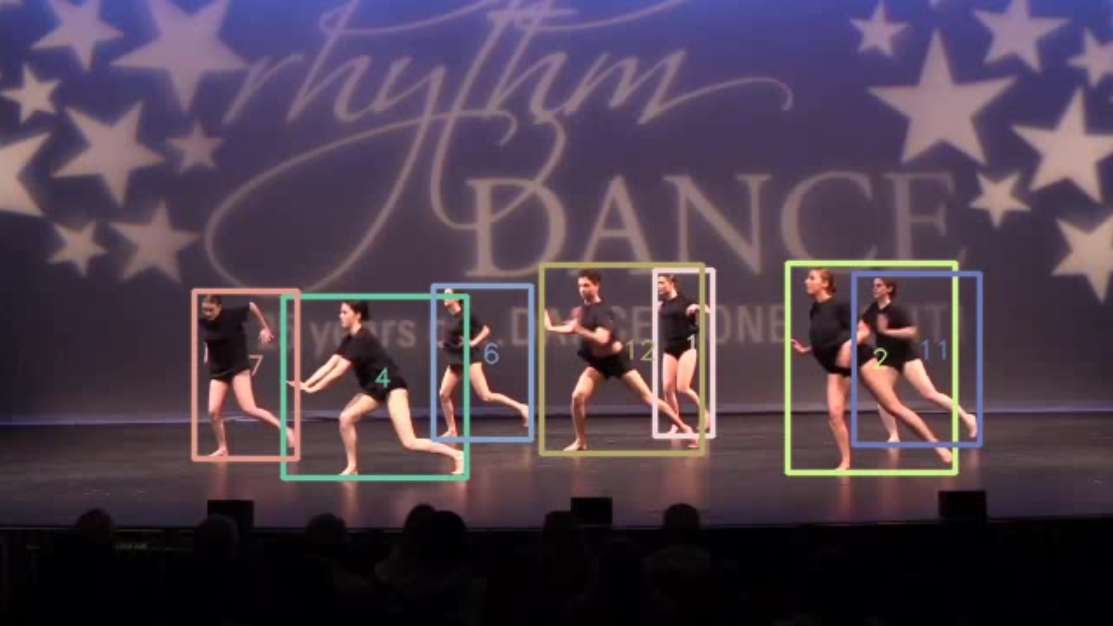}
\includegraphics[width=.31\linewidth]{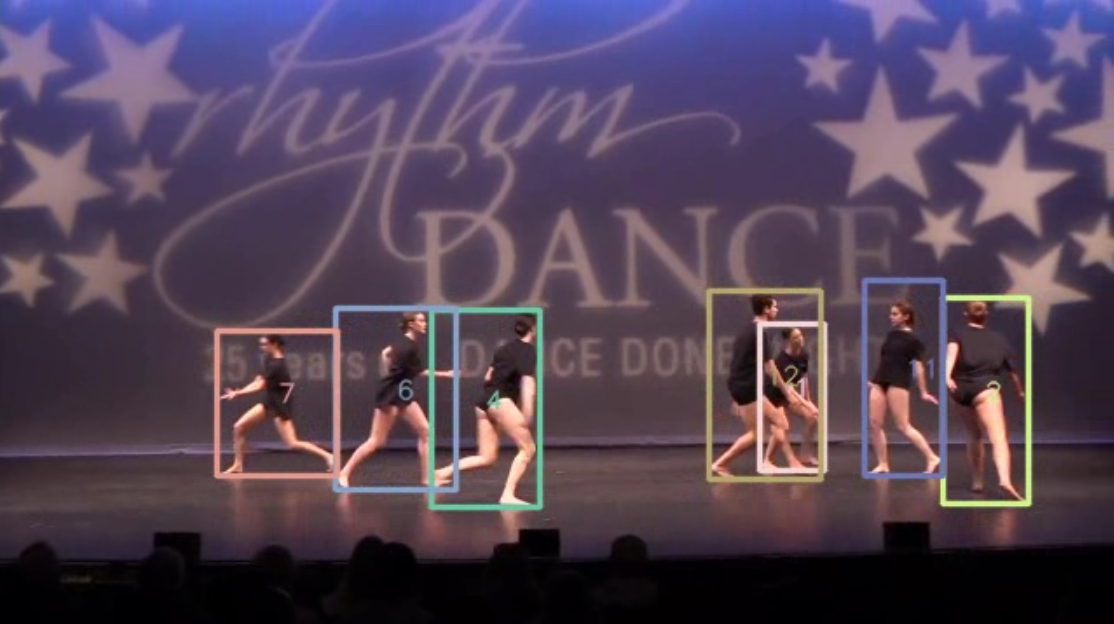}
\\
\vspace{3px}
\includegraphics[width=.31\linewidth]{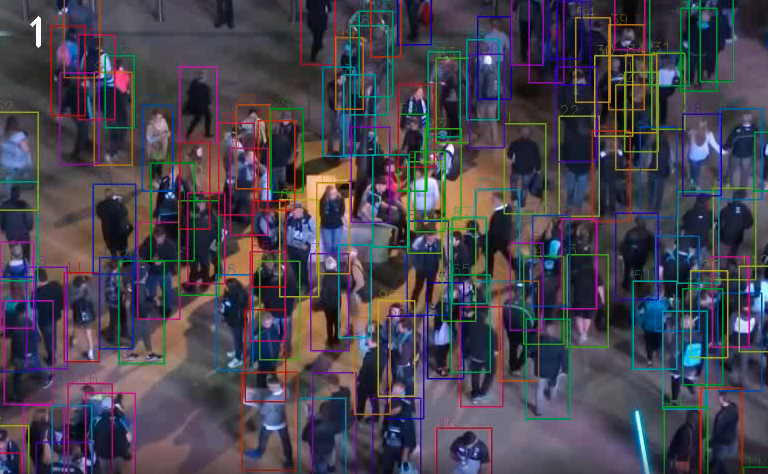}
\includegraphics[width=.31\linewidth]{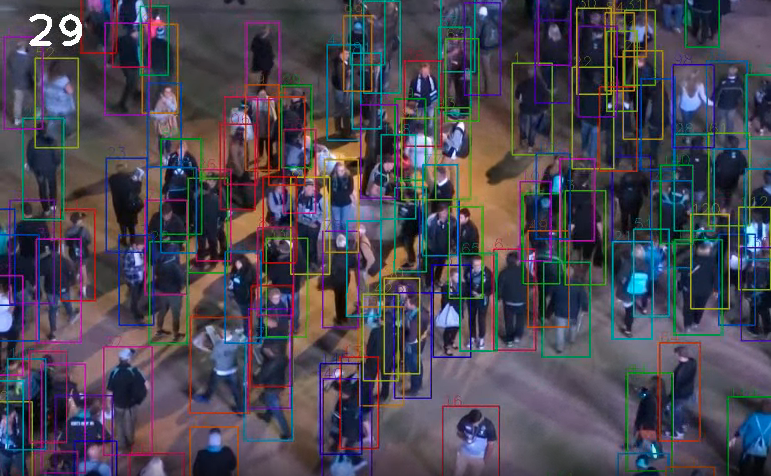}
\includegraphics[width=.31\linewidth]{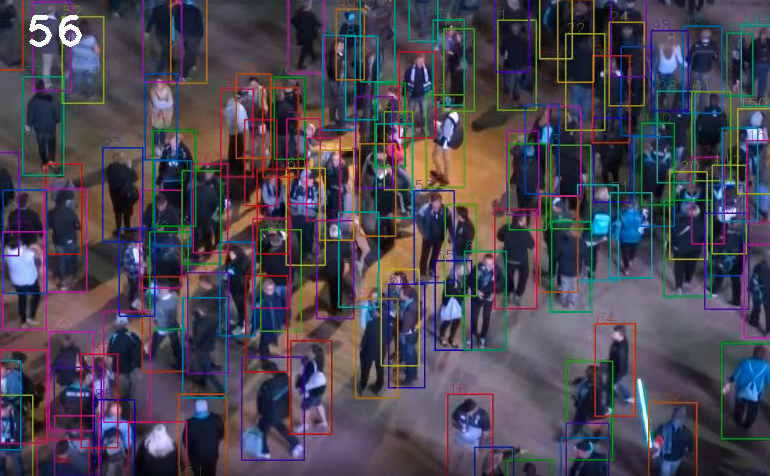}
\caption{\small \textcolor{red}{Upper line}: Results from DanceTrack-test set where targets have occlusion, crossover and similar appearance. \textcolor{red}{Bottom line}: Results on a MOT20-test video where the pedestrians are in the crowd and heavily occluded.}
\label{fig:dance_sample}
\vspace{-0.6cm}
\end{figure}

\subsection{Ablation Study} 
We now ablate the contribution of key variables in the design and implementation to the performance of \ours. 
Many previous works in the multi-object tracking community follow the practice of CenterTrack~\cite{centertrack} on MOT17~\cite{milan2016mot16} to use the latter half of training video sequences as the validation set. However, this makes the ablation study on the validation set not fair because the data distribution of the training set and validation set is so close that the performance gap reflected on the validation set might degrade or even disappear on the test set.
Therefore, we turn to DanceTrack~\cite{sun2021dancetrack} for the ablation study as an independent validation set is provided. For the following tables, we highlight our default implementation choice in \colorbox{yellow!30}{yellow}, which corresponds to the entries previously reported on benchmarks to compare with other methods.

\begin{table}[]
\centering
\scriptsize
    \caption{\small 
    Ablation of video clip length for training.}
    \begin{tabular}[t]{l | c c  c  c  c} \toprule
        $T$ & HOTA$\uparrow$ & DetA$\uparrow$ & AssA$\uparrow$ & MOTA$\uparrow$ & IDF1$\uparrow$ \\
        \midrule
        6 & 51.0 & 70.7 & 33.4 & 81.4 & 51.4\\ 
        \rowcolor{yellow!30} 8 & 51.9 & 71.4 & 34.0 & 81.9 & \textbf{52.2}\\ 
        10 & 52.4 & 71.7 & 34.5 & 81.8 & 51.4\\
        12 & \textbf{52.6} & \textbf{71.9} & \textbf{34.7} & \textbf{82.0} & 51.7\\
    \bottomrule
    \end{tabular}
    \label{tab:train_len}
    \vspace{-0.3cm}
\end{table}

\begin{table}[]
\centering
\scriptsize
    \caption{\small Ablation of video clip length for Inference.}
    \begin{tabular}[t]{l| c c  c c  c} \toprule
        $T$ & HOTA$\uparrow$ & DetA$\uparrow$ & AssA$\uparrow$ & MOTA$\uparrow$ & IDF1$\uparrow$ \\
        \midrule
        8 & 50.2 & 70.7 & 32.9 & 81.1 & 51.2\\
        16 & 51.6 & 71.2 & 33.6 & 81.5 & 51.7 \\
        \rowcolor{yellow!30} 24 &  \textbf{51.9} & \textbf{71.4} & \textbf{34.0} & 81.9 & \textbf{52.2}\\
        32 &  51.7 & 71.2 & 33.9 & \textbf{82.0} & 51.9\\
    \bottomrule
    \end{tabular}
    \label{tab:infer_len}
    \vspace{-0.3cm}
\end{table}

\begin{table*}[!htb]
\centering
\scriptsize
\caption{\small The ablation study about the contribution from \textit{semantic}, \textit{compositional}, and \textit{contextual} features.}
\setlength{\tabcolsep}{7pt}
\begin{tabular}{ p{20px}p{16px}p{20px} |ccccc}
\toprule
 Semantic & Compo. & Context. &  HOTA$\uparrow$ & DetA$\uparrow$ & AssA$\uparrow$ & MOTA$\uparrow$ & IDF1$\uparrow$\\
\midrule
\checkmark & &  & 47.8 & 69.1 & 30.1 & 80.8 & 49.1\\ 
\checkmark & \checkmark & & 49.6 & 69.3 & 31.3 & 81.2 & 50.4 \\ 
\checkmark & & \checkmark & 50.5 & 70.6 & 32.6 & 81.5 & 51.2 \\ 
\rowcolor{yellow!30} \checkmark & \checkmark & \checkmark & 51.9 & 71.4 & 34.0 & 81.9 & 52.2\\
\bottomrule
\end{tabular}
\label{table:module_ablation}
\vspace{-0.3cm}
\end{table*}

\begin{table*}[!htb]
\centering
\scriptsize
\caption{\small Different implementation choices to fit multiple training device configurations.}
\setlength{\tabcolsep}{7pt}
\begin{tabular}{ l| l|l|ccccc}
\toprule
 Training Device  & Train\_len & Image Size & HOTA$\uparrow$ & DetA$\uparrow$ & AssA$\uparrow$ & MOTA$\uparrow$ & IDF1$\uparrow$ \\
\midrule
 1x RTX 3090-24GB & 6 & 1280 $\times$ 1280 & 50.9 & 71.0 & 33.3 & 81.3 & 51.2 \\ 
 1x V100-32GB & 8 & 1560 $\times$ 1560 & 51.2 & 71.7 & 33.7 & 82.0 & 52.0\\  
 \rowcolor{yellow!30}4x V100-32GB & 8 & 1280 $\times$ 1280 & 51.9 & 71.4 & 34.0 & 81.9 & 52.2 \\
\bottomrule
\end{tabular}
\label{table:param_config}
\vspace{-0.3cm}
\end{table*}

\begin{table}[]
\centering
\scriptsize
    \caption{\small 
    Ablation of detector models.}
    \begin{tabular}[t]{l | p{13px}p{13px}p{13px}p{13px}p{13px} } \toprule
        Detector & HOTA$\uparrow$ & DetA$\uparrow$ & AssA$\uparrow$ & MOTA$\uparrow$ & IDF1$\uparrow$\\
        \midrule
        \rowcolor{yellow!30} CenterNet & 51.9 & 71.4 & 34.0 & 81.9 & 52.2\\ 
        YOLOv4~\cite{yolov4} & 52.6 & 73.8 & 34.5 & 84.0 & 53.4\\
        YOLOX~\cite{ge2021yolox} & \textbf{53.5} & \textbf{74.7} & \textbf{35.1} & \textbf{85.1} & \textbf{54.7}\\ 
    \bottomrule
    \end{tabular}
    \label{tab:ablation_detector}\vspace{-0.3cm}
\end{table}

\begin{table}[]
    \centering
\scriptsize
    \caption{\small Ablation about feature fusion strategies.}
    \begin{tabular}[t]{p{80px}|p{13px}p{13px}p{13px}p{13px}p{13px} } 
    \toprule
        Method & HOTA$\uparrow$ & DetA$\uparrow$ & AssA$\uparrow$ & MOTA$\uparrow$ & IDF1$\uparrow$\\
        \midrule
        Bbox only & 47.8 & 69.1 & 30.1 & 80.8 & 49.1\\ 
        Multi-Region CNN\cite{multiregioncnn} & 47.4 & 69.5 & 29.5 & 80.8 & 48.6\\ 
        \rowcolor{yellow!30} CSC-Attention & \textbf{51.9} & \textbf{71.4} & \textbf{34.0} & \textbf{81.9} & \textbf{52.2}\\ 
    \bottomrule
    \end{tabular}
    \label{tab:ablation_fuse}
    \vspace{-0.3cm}
\end{table}

\noindent\textbf{Video Length.} Table~\ref{tab:train_len} and~\ref{tab:infer_len} show the influence of video clip length in the training and inference stages respectively. The result suggests that training the association model with longer video clips can continuously improve performance. Limited by the GPU memory, we cannot increase the video clip length to longer than 12 frames here. On the contrary, during the inference stage, the sliding window size does not have a significant impact on the performance. Increasing the window size beyond a plateau will even hurt the performance.

\noindent\textbf{Three levels in CSC-hierarchy.} We study the contribution of each level of the CSC hierarchy in Table~\ref{table:module_ablation}. Here, only the semantic information is necessary for the evaluation with bounding box-based ground truth annotations and we can manipulate the other two levels in the CSC-hierarchy by not adding the corresponding feature in the generation of the CSC-Tokens.
Here we note that adding the compositional and contextual features only brings subtle computation overhead as the required self-attention and cross-attention operation are highly in parallel.
Compared to only using the \textit{semantic} feature, \ours achieves a significant performance improvement indicated by higher HOTA and AssA scores.
Also, integrating the features of the union area shows better effectiveness than solely integrating the features of body parts. 
This is probably because the cross attention between object body and union areas can provide critical information to compare object targets with their neighboring objects, preventing potential mismatch.
On the other hand, integrating the body part features can't explicitly avoid the mismatch with other instances. Fusing the features from all the levels turns out the best choice. 

\noindent\textbf{Input size.} We try different parameter configurations in Table~\ref{table:param_config} for the input clip length and image size. With only a single RTX 3090 GPU for training and inference, its performance is still comparable to the default setting with 4 $\times$ V100 GPUs. This makes the notorious computation barrier of transformer-based methods not that terrible anymore.

\noindent\textbf{Detector.} The highest priority for experiments is to validate the effectiveness of our proposed representations instead of racing on the leaderboard. For a fair comparison with the closest baseline GTR~\cite{gtr}, we follow it to choose CenterNet~\cite{centernet} as the default detector. But \ours is a tracking-by-detection method, flexible to integrate with different detectors. We compare CenterNet with the other detectors, i.e., YOLOv4~\cite{yolov4} and YOLOX~\cite{ge2021yolox} (used by ByteTrack, OC-SORT, SUSHI, etc.) in Table~\ref{tab:ablation_detector}. Advanced detectors can boost tracking performance.

\noindent\textbf{Fusion strategy of hierarchical features.} As a main contribution of this paper, we propose CSC-Attention module to fuse the features from the CSC-hierarchy. In a naive fashion, the multi-region CNN applies a \textit{split-and-concatenate} strategy to fuse the features from different bins inside a bounding box. We conduct a comparison with the multi-region CNN~\cite{multiregioncnn} in Table~\ref{tab:ablation_fuse}. Though multi-region CNN achieves improvement over the naive bounding box representation for object detection, this advantage is not observed anymore for multi-object tracking. Its performance gap with the features fused by CSC-Attention is even more significant than solely using the bounding box. This experiment suggests the effectiveness of the proposed three-level hierarchy and fusing them with the proposed CSC-Attention module.

\subsection{Robustness to Detection Noise}
With the enforcement of the part region (compositional) features, we expect \ours to show better robustness to the noise in detections. The intuition is that even if the bounding box is not accurate, as long as a distinct part is recognized, the model should be able to track an object consistently. To validate it, we add noise to the detection positions and observe its influence on the tracking performance. We apply random shifting and random resizing to add noise. For random shifting, we have a 25\% chance to shift the bounding box to the four directions independently, the shift stride is a random value in the range of $[0, \text{min}(0.2d, 20)]$, where $d$ is the bounding box width or height. We resize the bounding box width or height independently with a ratio of $\alpha_w$ and $\alpha_h$, both of which are random values in the range of [0.9, 1.1]. The results on Dancetrack-val are shown in Table~\ref{tab:robust}. Compared to the motion-based baseline OC-SORT and the full-box-only baseline GTR, \ours shows better robustness to the noise of detections as expected. 

\begin{table}[]
\centering
\scriptsize
    \caption{\small 
    Effect of detection noise (* indicates adding noise).}
    \begin{tabular}[t]{l |lll } \toprule
        Method & HOTA$\uparrow$  & AssA$\uparrow$ & 
        IDF1$\uparrow$ \\
        \midrule
        OC-SORT~\cite{ocsort} & 52.1	& 35.3	& 51.6\\ 
        OC-SORT* &  49.5 (\textcolor{brightturquoise}{$\downarrow$ 2.6})	& 31.3 (\textcolor{brightturquoise}{$\downarrow$ 4.0})& 48.5 (\textcolor{brightturquoise}{$\downarrow$ 3.1})\\ 
        GTR~\cite{gtr} & 47.2&	28.2&47.0\\ 
        GTR* &  45.0	(\textcolor{brightturquoise}{$\downarrow$ 2.2})& 26.7 (\textcolor{brightturquoise}{$\downarrow$ 1.5})& 45.6 (\textcolor{brightturquoise}{$\downarrow$ 1.4})\\ 
         \rowcolor{yellow!30} \ours & 51.9 & 34.0 & 	52.2\\ 
        \ours* & 50.8 (\textcolor{brightturquoise}{$\downarrow$ 1.1}) & 33.2 (\textcolor{brightturquoise}{$\downarrow$ 0.8}) & 51.5 (\textcolor{brightturquoise}{$\downarrow$ 0.7}) \\
    \bottomrule
    \end{tabular}
    \label{tab:robust}
\end{table}

\begin{table}[]
\centering
\scriptsize
    \caption{\small Time efficiency (MOT17-test).}
    \begin{tabular}[t]{l|ccc  } 
    \toprule
        Method & HOTA & training time & inference speed \\
        \midrule
        Transtrack~\cite{sun2020transtrack} & 54.1 & 18 hrs & 10FPS\\
        Trackformer~\cite{meinhardt2021trackformer} & - & - & 7.4FPS\\ 
        MOTR~\cite{zeng2021motr} & 57.2 & 63 hrs & 6.5FPS\\ 
        TransCenter~\cite{transcenter} & 54.5 & - & 11FPS\\ 
        GTR~\cite{gtr} & 59.1 & 4 hrs & 22.4FPS\\ 
        \rowcolor{yellow!30}  \ours & 60.8 & 4 hrs & 21.3FPS\\
    \bottomrule
    \end{tabular}
    \label{tab:time_efficiency}
    \vspace{-0.5cm}
\end{table}

\subsection{Time Efficiency}
Time efficiency is a bottleneck of query-based methods, especially for those using graph network~\cite{chu2021transmot}, long-history buffers~\cite{memot} or temporal aggregation~\cite{zeng2021motr}. Collecting the methods that report the time efficiency or have open-sourced implementation, we report the required training time and inference speed in Table~\ref{tab:time_efficiency} by default settings on MOT17. The speed is tested on Nvidia V100 GPU and the training time is evaluated on 4xV100 GPUs. \ours achieves the best accuracy with one of the best time efficiency for both training time and the inference speed.

\section{Conclusion}
In this paper, we propose to construct discriminative visual representations by a \textit{compositional-semantic-contextual} visual hierarchy combining different visual cues to distinguish a target.
To leverage them comprehensively, we propose a CSC-Attention to gather and fuse the visual features. These are the two main contributions of this paper. We have demonstrated that they are connected to show power. The designs are integrated into \ours for multi-object tracking. The results on multiple datasets demonstrate its efficiency and effectiveness. We hope the study of this paper can provide new knowledge in the visual representation of objects and an advanced baseline model to solve multi-object tracking problems. The method is also more robust to the detection noises and computation-economic.

\bibliography{ref}
\bibliographystyle{IEEEtranS}

\end{document}